\documentclass{article}

\usepackage{microtype}
\usepackage{graphicx}
\usepackage{booktabs} 

\usepackage{hyperref}

\usepackage[accepted]{icml2023}

\usepackage{amsmath}
\usepackage{amssymb}
\usepackage{mathtools}
\usepackage{amsthm}
\usepackage{comment}
\usepackage{hyperref}       %
\usepackage{url}            %
\usepackage{booktabs}       %
\usepackage{amsfonts}       %
\usepackage{nicefrac}       %
\usepackage{microtype}      %
\usepackage{color}
\usepackage{amsmath}
\usepackage{amsthm}
\usepackage{graphicx}

\usepackage{comment}
\usepackage{tikz}
\usepackage{natbib}
\usepackage{caption}
\usepackage{multirow}
\usepackage{wrapfig}
\usepackage{amsfonts, amsmath}
\usepackage{enumitem}

\usepackage{varwidth}
\usepackage{lipsum}
\usepackage{multicol}
\newsavebox{\algbox}
\usepackage{import}
\usetikzlibrary{bayesnet}
\usetikzlibrary{arrows}

\usepackage[utf8]{inputenc} 
\usepackage[T1]{fontenc}    
\usepackage{hyperref}       
\usepackage{url}            
\usepackage{booktabs}       
\usepackage{amsfonts}       
\usepackage{nicefrac}       
\usepackage{microtype}      
\usepackage{xcolor}         
\usepackage{url}            %
\usepackage{booktabs}       %
\usepackage{amsfonts}       %
\usepackage{nicefrac}       %
\usepackage{microtype}      %
\usepackage{color}
\usepackage{amsmath}
\usepackage{amsthm}
\usepackage{graphicx}
\usepackage{comment}
\usepackage{tikz}
\usepackage{natbib}
\usepackage{caption}
\usepackage{multirow}
\usepackage{wrapfig}
\usepackage{amsfonts, amsmath}
\usepackage{enumitem}
\usepackage{soul}
\usepackage{bm}
\usepackage{wrapfig}
\usepackage{listings}
\usepackage[title]{appendix}
\usepackage[bottom]{footmisc}
\usepackage[makeroom]{cancel}
\usepackage[normalem]{ulem}
\setcitestyle{authoryear,open={(},close={)}}
\usepackage{array}
\newcolumntype{P}[1]{>{\centering\arraybackslash}p{#1}}
\usepackage{varwidth}
\usepackage{lipsum}
\usepackage{multicol}
\usepackage{import}
\usetikzlibrary{bayesnet}
\usetikzlibrary{arrows}
\usepackage{booktabs}

\usepackage[utf8]{inputenc} 
\usepackage[T1]{fontenc}    
\usepackage{hyperref}       
\usepackage{url}            
\usepackage{booktabs}       
\usepackage{amsfonts}       
\usepackage{nicefrac}       
\usepackage{microtype}      
\usepackage{xcolor}         
\def\abode{\texttt{AbODE}}

\def\calL{\mathcal{L}}
\def\x{\mathbf{x}}
\def\s{\mathbf{s}}
\def\a{\mathbf{a}}

\def\u{\mathbf{u}}
\def\dz{\dot{\mathbf{z}}}

\def\z{\mathbf{z}}

\usepackage[capitalize,noabbrev]{cleveref}

\theoremstyle{plain}

\theoremstyle{definition}

\theoremstyle{remark}

\usepackage[textsize=tiny]{todonotes}

\icmltitlerunning{AbODE: Ab Initio Antibody Design using Conjoined ODEs}

\begin{document}

\twocolumn[
\icmltitle{AbODE: Ab Initio Antibody Design using Conjoined ODEs}



\icmlsetsymbol{equal}{*}

\begin{icmlauthorlist}
\icmlauthor{Yogesh Verma}{yyy}
\icmlauthor{Markus Heinonen}{yyy}
\icmlauthor{Vikas Garg}{yyy,comp}
\end{icmlauthorlist}

\icmlaffiliation{yyy}{Department of Computer Science, Aalto University, Finland}
\icmlaffiliation{comp}{YaiYai Ltd}

\icmlcorrespondingauthor{Yogesh Verma}{yogesh.verma@aalto.fi}

\icmlkeywords{Machine Learning, ICML}

\vskip 0.3in
]



\printAffiliationsAndNotice{} 

\begin{abstract}
Antibodies are Y-shaped proteins that neutralize pathogens and constitute the core of our adaptive immune system. {\em De novo} generation of new antibodies that target specific {\em antigens} holds the key to accelerating vaccine discovery. However, this co-design of the amino acid sequence and the 3D structure subsumes and accentuates, some central challenges from multiple tasks including  protein folding (sequence to structure), inverse folding (structure to sequence), and docking (binding). 

We strive to surmount these challenges with a new generative model \abode~that extends graph PDEs to accommodate both contextual information and external interactions. Unlike existing approaches, \abode~ uses a single round of full-shot decoding, and elicits continuous differential attention that encapsulates, and evolves with, latent interactions within the antibody as well as those involving the antigen. We unravel fundamental connections between \abode~ and temporal networks as well as graph-matching networks. The proposed model significantly outperforms existing methods on standard metrics across benchmarks.
\end{abstract}

\section{Introduction}
\label{intro}
Machine learning methods have recently enabled exciting developments for computational drug design, including, on tasks such as protein folding \cite{pf4}, i.e., predicting the 3D structure of a given protein from its amino acid sequence; 
sequence design or inverse folding \cite{ingraham2019generative}, i.e., generating new sequences that fold into a given 3D structure; and docking \cite{equidock}, i.e., predicting the complex when two proteins bind together. 

\begin{figure*}[!ht]
\vskip 0.2in
\begin{center}
\centerline{\includegraphics[width=\textwidth]{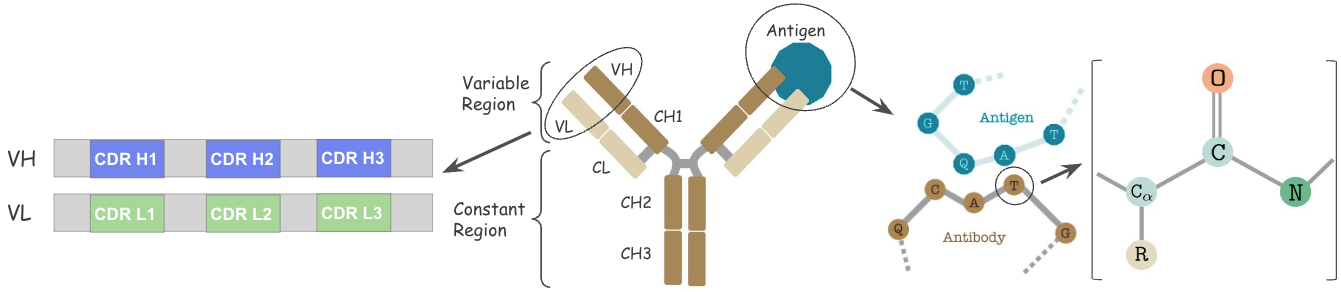}}
\caption{Schematic showing the structure of a residue (amino acid), where the backbone atoms we use are $N$, $C_{\alpha}$ and $C$ (\textbf{right}) and the structure of the antibody (\textbf{left}) which is Y-shaped showing the VH/VL sequences and binding to the antigen, and we focus on CDRs of the variable domain in the heavy chain (VH).}
\label{fig:ab}
\end{center}
\vskip -0.2in
\end{figure*}

We focus on the problem of {\em antibody design}. Antibodies, the versatile Y-shaped proteins that guard against pathogens such as bacteria and viruses, are essential to our adaptive immune mechanism. Typically, an antibody acts by binding to a specific molecule of the pathogen, namely, the {\em antigen}. Each antibody recognizes a unique antigen, and the so-called Complementarity Determining Regions (CDRs) at the tip of the antibody determines this specificity (\autoref{fig:ab}). Thus, automating the design of antibodies against specific pathogens (e.g., the SARS-CoV-2 virus) can revolutionize drug discovery \cite{pinto2020, refinegnn}.  

Our objective is to co-design the CDR sequence and structure from scratch, conditioned on an antigen. However, significant challenges must be overcome in this pursuit. While recent generative methods for protein sequence design have been successful \cite{ingraham2019generative}, they crucially utilize that the long term dependencies in sequence are local in the 3D space. However, the CDR structures are seldom known a priori, thereby limiting the scope of such approaches \cite{refinegnn}. In principle, one could segregate the design of sequence from structure. Indeed, once a CDR sequence is generated, folding methods such as AlphaFold that exploit alignment with a family of protein sequences \cite{pf4} can be employed to estimate the 3D structure of the CDR. However, generating sequence without conditioning on the structure \citep{prot2,antgen1} is known to produce sub-optimal sequences. Moreover, related sequences may be unavailable for scenarios that diverge considerably from naturally occurring antibodies \cite{ingraham2019generative}. Finally, finding antibodies that have a good binding affinity with the target antigens \cite{Raybould} requires search in a huge space ($\sim 20^{60}$ possible CDR sequences).         

Initial approaches for antibody design \citep{ant1,ant2,ant3,rabd} relied on hand-crafted energy functions that entailed expensive simulation, and could not sufficiently capture complex interactions ~\citep{ant4}. Going beyond 1D sequence prediction \citep{prot2,antgen1,antgen2,akbar2022silico},  recent generative methods co-design structure and sequence \cite{refinegnn} and can incorporate information about antigen directly in the model \cite{pmlr-v162-jin22a, mean}. 
 
Certain shortcomings, however, accompany these advances. The autoregressive scheme (one residue at a time) adopted by \cite{pmlr-v162-jin22a, refinegnn} is susceptible to issues such as vanishing or exploding gradients during training, as well as slow generation and accumulation of errors during inference. \citet{mean} advocate multiple {\em full-shot} rounds to address this issue; however, segregating context (intra-antibody) from external interactions (antibody-antigen) precludes joint optimization, and may result in sub-optimality.   

We circumvent these issues with a novel viewpoint that models the antibody-antigen complex as a joint 3D graph with heterogeneous edges. Different from all prior works, this perspective allows us to formulate a coupled neural ODE system over the nodes pertaining to the antibody, while simultaneously accounting for the antigen. Specifically, we associate local densities (one per antibody node) that are progressively refined toward globally aligned densities based on simultaneous feedback from the antigen as well as the (other) antibody nodes. The 3D coordinates and the node labels for the antibody can then be sampled after a few rounds in {\em one-shot}, i.e., all at once. Thus, the entire procedure is efficient and end-to-end trainable. 

We show how invariance can be built in readily into the proposed method \abode~ toward representations that account for rotations and other symmetries. \abode~ establishes a new state-of-the-art (SOTA) for antibody design across standard metrics on several benchmarks. Interestingly, it turns out that it shares  connections with two recent methods for equivariant molecular generation and docking, namely, ModFlow and IEGMN. While ModFlow can be recovered as a special case of the \abode~ formulation, IEGMN may be interpreted as a discrete analog of \abode. One one hand, these similarities reaffirm the kinship of different computational drug design tasks; on the other, they suggest the broader applicability of neural PDEs as effective tools for these tasks. Our experiments further reinforce this phenomenon: \abode~ is already competitive with the SOTA methods on a task it is not tailored for, namely, fixed backbone protein sequence design.

\subsection{Contributions} \label{contributions} 
 In summary, we make following contributions. 

\begin{itemize}
    \item We propose \abode, a generative model that extends graph PDEs by jointly modeling the internal context and interactions with external objects (e.g., antigens). 
    \item \abode~ co-designs the antibody sequence and structure, 
    using a single round of full-shot decoding.
    \item Empirically, ~\abode~ registers SOTA performance on various sequence design and structure prediction tasks.
\end{itemize}
\begin{figure*}[!tbp]
\vskip 0.1in
\begin{center}
\centerline{\includegraphics[width=\textwidth]{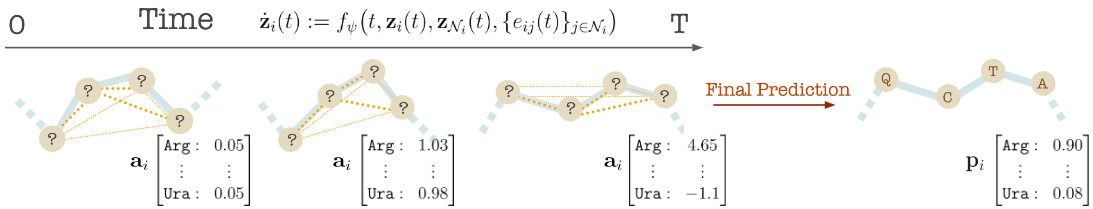}}
\caption{A demonstration of \abode. The initial structure and amino acid labels  evolve in time under $f_{\psi}$ and are subsequently transformed into a final structure and amino acid labels.}
\label{fig:work}
\end{center}
\vskip -0.1in
\end{figure*}
\section{Related Work}
\label{related}

\paragraph{Antibody/protein design} Early approaches for computational antibody design 
optimize hand-crafted energy functions
\citep{ant1,ant2,ant3,rabd}. These methods require costly simulations and are prone to defects due to complex interactions between chains that cannot be captured by force fields or statistical functions~\citep{ant4}. Recently, deep generative models have been utilized for 1D sequence prediction in proteins \citep{prot1,ingraham2019generative,prot3,prot4,prot5,protmpnn} and antibodies \citep{prot2,antgen1,antgen2,akbar2022silico}, conditioned on the backbone 3D structure. \citet{refinegnn} proposed to co-design the sequence and structure via an autoregressive refinement technique, while \citet{mean} advocated multiple rounds of full-shot decoding together with an encoder for intra-antibody context, and a separate encoder for external interactions. 

Different from all these works, we formulate a single full-shot method that extends graph PDEs\citep{chamberlain2021grand,iakovlev2020learning} to accommodate and condition on spatial and context-based information of the antigen, tailored to antibody sequence and structure generation.

\color{black}

\paragraph{Generative models for graphs} Our work is related to continuous time models for graph generation \citep{modflow,avelar2019discrete} that incorporate dynamic interactions~\citep{chen2018neural,grathwohl2018ffjord,iakovlev2020learning,eliasof2021pde} over graphs. Methods have also been developed for protein structure generation, e.g., Folding Diffusion~\citep{folding}, \citet{anand2018}, AlphaDesign~\citep{alphadesign}, etc. Most of these methods lack the flexibility to be directly applied to antibody sequence and structure design, due to their inability to capture effective inductive biases, conditional information, and higher-order features. In contrast, we can combine conditional information and evolve the structure and sequence via latent co-interacting trajectories. 

\paragraph{3D structure prediction} Our method is also closely related to docking ~\citep{equidock,equibind} and protein folding~\citep{pf1,pf2,pf3,pf4,pf5}. Methods like DiffDock~\citep{diffdock} and EquiBind~\cite{equibind} predict only the structure of the molecule given a protein binding site but lack any generative component related to sequence design. AlphaFold~\citep{pf4} requires holistic information like  protein sequence, multi-sequence alignment (MSA), and template features. These models cannot be directly applied for antibody design, where MSA is not specified in advance and one needs to predict the structure of an incomplete sequence. In contrast, we learn to co-model the 3D structure and sequence for incomplete graphs and interleave structure modeling with sequence prediction.  


\section{Antibody sequence and structure co-design}

An antibody (Ab) is a Y-shaped protein (Fig.~\ref{fig:ab}) that identifies antigens of a foreign object (e.g., a virus) and stimulates an immunological response. An antibody consists of a constant domain, and a symmetric variable region divided into heavy (H) and light (L) chains \cite{kuroda}. The surface of the antibody contains three complementarity-determining regions (CDRs), which act as the main binding determinant. CDR-H3 makes up the majority of the binding affinity \cite{fischman2018computational}. The non-CDR regions are highly preserved \cite{kuroda}; thus, it is common to formulate antibody design as a CDR design problem \cite{shin2021protein}. 

We view the antibody-antigen complex as a joint graph with interactions between nodes across the binding. We co-model both the sequence and the 3D conformation of the CDR regions with a graph PDE and apply our method to antigen-specific and unconditional antibody design tasks. 

We seek a representation that is invariant to translations and rotations due to its locality along the backbone. Moreover, we would like the edge features to be sufficiently informative such that the relative neighborhoods can be reconstructed up to rigid body motion \cite{ingraham2019generative}. We describe next a representation that satisfies these desiderata.

\subsection{The antibody-antigen graph}
We define the antigen-antibody complex as a 3D graph $G=(V, E, X)$, with antibody $\texttt{Ab}$ and antigen $\texttt{Ag}$ vertices $V = (V_{\texttt{Ab}}, V_{\texttt{Ag}} )$, coordinates $X = (X_{\texttt{Ab}}, X_{\texttt{Ag}} )$ and edges $E = (E_{\texttt{Ab}}, E_{\texttt{Ab-Ag}} )$ within the antibody as well as between the antibody and the antigen. Each vertex $v \in \mathcal{A} = \{ \texttt{Arg},\texttt{His},\ldots \}$ is one of 20 amino acids. We treat the labels with a Categorical distribution, such that the label features $\a_{i} \in \mathrm{R}^{20}$ represent the unnormalized amino acid probabilities. We also represent each residue by the cartesian 3D coordinates of its three backbone atoms $\{ N, C_{\alpha}, C\}$ (see Fig.~\ref{fig:ab}). For the $i^{th}$ residue $\x_{i}$ we compute its spatial features $\s_{i} = (r_{i},\alpha_{i},\gamma_{i})$ in Eq.~\ref{eq:space}, where, $r_i$ denotes the distance between consecutive residues $x_i$ and $x_{i+1}$, $\alpha_{i}$ is the co-angle of residue $i$ wrt previous and next residue, $\gamma_{i}$ is the azimuthal angle of $i$’s local plane, and $\mathbf{n}_{i}$ is the normal vector. The full residue state $\z_i = [\a_i, \s_i]$ concatenates the label features $\a_i$ and the spatial features $\s_i$.
\begin{align} \label{eq:space}
r_i & =\left\|\mathbf{u}_i\right\|, \quad \mathbf{u}_i=\mathbf{x}_{i+1}-\mathbf{x}_i \\ \alpha_i & =\cos ^{-1}\left(\frac{\left\langle\mathbf{u}_i, \mathbf{u}_{i-1}\right\rangle}{\left\|\mathbf{u}_i\right\| \cdot\left\|\mathbf{u}_{i-1}\right\|}\right) \\ \gamma_i & =\cos ^{-1}\left(\frac{\left\langle\mathbf{u}_i, \mathbf{n}_i\right\rangle}{\left\|\mathbf{u}_i\right\| \cdot\left\|\mathbf{n}_i\right\|}\right), \quad \mathbf{n}_i=\mathbf{u}_i \times \mathbf{u}_{i-1} .
\end{align}

\begin{figure}[!htb]
    \centering
    \includegraphics[scale=0.4]{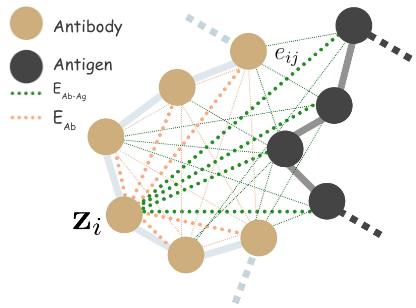}
    \caption{Schematic graph construction for the antigen-antibody complex with internal edges $E_{\texttt{Ab}}$ and external edges $E_{\texttt{Ab-Ag}}$. In the unconditional setting (i.e., the antigen is not specified), this reduces to an antibody graph}
    \label{fig:ab_ag_comp}
\end{figure}

\paragraph{Interactions} To capture the interactions pertaining to the complex, we define edges $E_{\texttt{Ab}}$ between all antibody residues and edges $E_{\texttt{Ab-Ag}}$ between all antibody and antigen residues (See Figure~\ref{fig:ab_ag_comp}). We also define edge features between nodes $i$ and $j$,
\begin{align} \label{eq:edge_feat}
\mathbf{e}_{i j}=(&\Delta \mathbf{z}_{i j}, i-j, \operatorname{RBF}\left(\left\|\mathbf{s}_i-\mathbf{s}_j\right\|\right). \\ 
& \mathcal{O}_i^{\top} \frac{s_{i, \alpha}-s_{j, \alpha}}{\left\|s_{i, \alpha}-s_{j, \alpha}\right\|}, \mathcal{O}_i^{\top} \mathcal{O}_j, k_{i j}) .
\end{align}
These include state differences $\Delta \mathbf{z}_{i j} = \{ \Delta \a_{ij}, \Delta \s_{ij}\}$ over label features $\Delta \a_{ij} = \a_j-\a_i$ and spatial features $\Delta \s_{ij} = \{(\Delta r_{ij},\Delta \alpha_{ij},\Delta \gamma_{ij})_{p} \mid p \in \{ N,C_{\alpha},C\} \}$, backbone distance $i-j$, and spatial distance $\texttt{RBF}(||\s_i-\s_j||)$ (here, RBF is the standard radius basis function kernel). The fourth term encodes directional embedding in the relative direction of $j$ in the local coordinate frame $\mathcal{O}_i $ \cite{ingraham2019generative}, and the $\mathcal{O}^{T}_i\mathcal{O}_j $ describes the orientation encoding of the node $i$ with node $j$ (See Appendix~\ref{sec:orient} for details). Finally, we encode within-antibody edges with $k = 1$ and antibody-antigen edges with $k = 2$.
\paragraph{Task formulation} Given a three-dimensional antibody or antibody-antigen graph, we aim to learn a PDE in order to generate an amino acid sequence and the corresponding 3D conformation jointly.
\begin{table*}[!t]
    \vskip 0.1in
\centering
    \caption{\abode~ as a variant of Independent E(3)-Equivariant Graph Matching Network (IEGMN) applied to interactions among two graphs $G_{1} = (V_1,E_1)$ and $G_{2} = (V_2,E_2)$. Here, $e_{ij} \in E_1 \cup E_2$; $n \in V_1 \cup V_2$; $\texttt{RBF}(\x_i,\x_j;\sigma) = \texttt{exp}(-||\x_{i}^{(l)} - \x_{j}^{(l)}||^{2}/\sigma)$; $h_n$ and $x_n$ denote, respectively, the node embedding and the spatial embedding; $a_{ij}$ are attention based coefficients; $\phi^{x}$ is a real-valued (scalar) parametric function; $\phi^{h,e}$ are parametric functions (MLPs); $\textbf{f}_{ij},\textbf{f}_{i}$ are the original edge and node features; $\beta,\eta$ are scaling parameters and $\mathbf{W}$ is a learnable matrix. For \abode~, $\alpha_{i,j}$ are the attention coefficients; $\mathbf{W}_{1},\ldots,\mathbf{W}_{6}$ are learnable weight parameters; $d$ is the hidden size of each head; $\mathcal{N}_{int}(i)$ are the neighbours $j$ of node $i$ such that $k_{ij} = 1$, and $\mathcal{N}_{ext}(i)$ are the neighbours such that $k_{ij} = 2$.  }
    \label{tab:iegmn_tab}
    \centering
    \resizebox{\textwidth}{!}{
    \begin{tabular}{lcc}
        \toprule
        Method  & IEGMN layer &  \abode   \\
        \midrule
        
        \multirow{2}{4em}{Edge}  & $\mathbf{m}_{i j}=\varphi^e\left(\mathbf{h}_i^{(l)}, \mathbf{h}_j^{(l)}, \operatorname{RBF}\left(\mathbf{x}_i^{(l)}, \mathbf{x}_j^{(l)} ; \sigma\right), \mathbf{f}_{i j}\right)$ & $\alpha_{i, j}=\operatorname{softmax}\left(\frac{(\mathbf{W}_{3}\z_{i})^{\top}(\mathbf{W}_{4}\mathbf{z}_j+\mathbf{W}_{6}\mathbf{e}_{i,j}).}{\sqrt{d}}\right)$  \\
        
         & $\mathrm{m}_n=\frac{1}{|\mathcal{N}(n)|} \sum_{j \in \mathcal{N}(n)} \mathrm{m}_{n j}$  & $m_i^{\prime}=\sum_{j \in \mathcal{N}_i} \alpha_{i, j}\left(\mathbf{W}_2 \mathbf{z}_j+\mathbf{W}_6 \mathbf{e}_{i j}\right)$ \\
         \midrule
        Intra and Inter connections  & $\boldsymbol{\mu}_{i j}=a_{i j} \mathbf{W h}_j^{(l)}, \forall i \in \mathcal{V}_1, j \in \mathcal{V}_2 \text { or } i \in \mathcal{V}_2, j \in \mathcal{V}_1$  &$m_{ij}^{\prime,ext} = \alpha_{i,j}\left(\mathbf{W}_2 \mathbf{z}_j+\mathbf{W}_6 \mathbf{e}_{i j}\right),~m_{ij}^{\prime,int} = \alpha_{i,j}\left(\mathbf{W}_2 \mathbf{z}_j+\mathbf{W}_6 \mathbf{e}_{i j}\right)$   \\
        & $\boldsymbol{\mu}_i=\sum_{j \in \mathcal{V}_2} \boldsymbol{\mu}_{i j}, \forall i \in \mathcal{V}_1, \quad \boldsymbol{\mu}_k=\sum_{l \in \mathcal{V}_1} \boldsymbol{\mu}_{k l}, \forall k \in \mathcal{V}_2$  & $\textbf{m}_i^{\prime,int}=\sum_{j}^{\mathcal{N}_{int}(i)}m_{ij}^{\prime,int},~\textbf{m}_i^{\prime,ext}=\sum_{j}^{\mathcal{N}_{ext}(i)}m_{ij}^{\prime,ext}$ \\
        \midrule
        Node embedding & $\mathbf{h}_n^{(l+1)}=(1-\beta) \cdot \mathbf{h}_n^{(l)}+\beta \cdot \varphi^h\left(\mathbf{h}_n^{(l)}, \mathbf{m}_n, \boldsymbol{\mu}_n, \mathbf{f}_n\right)$ & $\mathrm{a}_i^{\prime}=\mathbf{W}_1 \mathbf{a}_i+\mathbf{m}_i^{\prime, \text{int}}+\mathbf{m}_i^{\prime,\text{ext}}$  \\
        \midrule
        Coordinate embedding & $\mathbf{x}_n^{(l+1)}=\eta \mathbf{x}_n^{(0)}+(1-\eta) \mathbf{x}_n^{(l)}+\sum_{j \in \mathcal{N}(n)}\left(\mathbf{x}_n^{(l)}-\mathbf{x}_j^{(l)}\right) \varphi^x\left(\mathbf{m}_{nj}\right)$ & $\mathrm{s}_i^{\prime}=\mathbf{W}_{1}\s_i+\mathbf{m}_i^{\prime,\text{int}}+\mathbf{m}_i^{\prime, \text{ext}}$  \\
        \bottomrule
    \end{tabular}}
    \vskip -0.1in
\end{table*}
\subsection{Conjoined system of ODEs}\label{sec:ode}
We propose to model the distribution of antibody-antigen complexes by a differential graph flow $\z(t)$ over time $t \in \mathrm{R}_{+}$. We initialize the initial state $\z(0)$ to a uniform categorical vector, similar to mask initialization \cite{refinegnn,mean}. Coordinates are initialized with the even distribution between the residue right before CDRs and the one right after CDRs following \cite{mean}, and we learn a differential $\frac{d\z(t)}{dt}$ that maps to the end state $\z(T)$ that matches data.

We begin by assuming an ODE system $\{\z_{i}(t)\}$ over time $t \in \mathrm{R}_{+}$, where node the time evolution of node $i$ is an ODE
\begin{align}
    \dz_{i}(t) = \frac{\partial \z_{i}(t)}{\partial t} = f_{\psi}(t,\z_{i}(t),\z_{N(i)}(t),\{\mathbf{e}_{ij}(t)\}_{j})
\end{align}
where $N(i) = \{ j~:~(i,j)~\in~E\}$  indexes the neighbors of node $i$, and the function $f$ parameterized by $\psi$ is our main learning goal. The differentials form a coupled ODE system
\begin{align} \label{eq:dz}
    \dz(t) &= \begin{pmatrix} \dz_1(t) \\ \vdots \\ \dz_M(t) \end{pmatrix}\\ 
    &= \begin{pmatrix} f_{\psi}(t,\z_{1}(t),\z_{N(1)}(t),\{\mathbf{e}_{1j}(t)\}_{j}) \\ \vdots \\ f_{\psi}(t,\z_{M}(t),\z_{N(M)}(t),\{\mathbf{e}_{Mj}(t)\}_{j}) \big) \end{pmatrix} \\
    \z(T) &= \z(0) + \int_0^T \dz(t) dt~.
\end{align}
where $M$ is the number of nodes. The above ODE system corresponds to a graph PDE \cite{iakovlev2020learning,modflow}, whose forward pass and backpropagation can be solved efficiently by ODE solvers.

Interestingly, it turns out that the PDE about a recently proposed method for molecular generation can be recovered as a particular case of \ref{eq:dz}, when all the edges are set to be of the same type.
\paragraph{Proposition 1}: \emph{ModFlow~\cite{modflow} can be seen as a special case of \abode~ in an unconditional setting. This can be achieved by setting $k_{ij}=1$ for every $e_{ij}$.}

\subsection{Attention-based differential}
We capture the interactions between the antigen and antibody residues with graph attention~\cite{shi2020masked}
\begin{align} \alpha_{i j} & =\operatorname{softmax}\left(\frac{\left(\mathbf{W}_3 \mathbf{z}_i\right)^{\top}\left(\mathbf{W}_4 \mathbf{z}_j+\mathbf{W}_6 \mathbf{e}_{i j}\right)}{\sqrt{d}}\right) \\ \mathbf{z}_i^{\prime} & =\mathbf{W}_1 \mathbf{z}_i+\sum_{j \in N(i)} \alpha_{i j}\left(\mathbf{W}_2 \mathbf{z}_j+\mathbf{W}_6 \mathbf{e}_{i j}\right)\end{align}
where  $\textbf{W}_1,\ldots,\textbf{W}_6$ are weight parameters and $d$ is the head size. The $\alpha$'s are the attention coefficients corresponding to within and across edges, which are used to update the node feature $\z_i$. Interestingly, our method also shares similarities with the Independent E(3)-Equivariant Graph Matching Networks (IEGMNs) for docking~\cite{equidock}.
\paragraph{Proposition 2}: \emph{\abode~ can be cast as Independent E(3)- Equivariant Graph Matching Networks (IEGMN)~\cite{equidock}). The operations are listed in Table~\ref{tab:iegmn_tab} (See Appendix~\ref{sec:iegmn} for more details).}

In this sense, our extended graph PDE unifies molecular generation and docking with protein/antibody design. 

We now describe our training objective.

\begin{table*}[!t]
\centering
\vskip 0.1in
\caption{\textbf{Top}: Unconditional sequence and structure benchmark. We report perplexity (PPL) and root mean square deviation (RMSD) for each CDR in the heavy chain. Baselines are from \citet{refinegnn}. \textbf{Bottom}: Antigen-conditional sequence and structure benchmark on SAbDab~\cite{dunbar2014sabdab}. We report amino acid recovery (AAR) and root mean square deviation (RMSD) for each CDR in the heavy chain. Baselines are from~\citet{mean}.}
\label{tab:result_cond_uncond}
\centering
\resizebox{0.89\textwidth}{!}{
\begin{tabular}{lcccccc}
\hline
& \multicolumn{2}{c}{CDR-H1} & \multicolumn{2}{c}{CDR-H2}  & \multicolumn{2}{c}{CDR-H3}   \\
\cmidrule(lr){2-3} \cmidrule(lr){4-5} \cmidrule(lr){6-7}
Method & PPL ($\downarrow$) & RMSD ($\downarrow$) & PPL ($\downarrow$) & RMSD ($\downarrow$)  & PPL ($\downarrow$) & RMSD ($\downarrow$)   \\
\hline
LSTM & $6.79$ & \textcolor{gray}{(N/A)} & $7.21$ & \textcolor{gray}{(N/A)} & $9.70$ & \textcolor{gray}{(N/A)} \\
AR-GNN & $6.44$ & $2.97$ & $6.86$ & $2.27$ & $9.44$ & $3.63$\\
RefineGNN & $6.09$ & $1.18$ & $6.58$ & $0.87$ & $8.38$ & $2.50$\\
\abode &  \textbf{$4.25$} $\pm$ \textcolor{gray}{$0.46$} & \textbf{$0.73$} $\pm$ \textcolor{gray}{$0.15$} & \textbf{$4.32$} $\pm$ \textcolor{gray}{$0.32$} &\textbf{$0.63$} $\pm$ \textcolor{gray}{$0.19$}   & \textbf{$6.35$} $\pm$ \textcolor{gray}{$0.29$}  & \textbf{$2.01$} $\pm$ \textcolor{gray}{$0.13$}  \\
\hline
\end{tabular}}
\vskip 0.15in
\resizebox{0.91\textwidth}{!}{
\begin{tabular}{lcccccc}
\hline
& \multicolumn{2}{c}{CDR-H1} & \multicolumn{2}{c}{CDR-H2}  & \multicolumn{2}{c}{CDR-H3}  \\
\cmidrule(lr){2-3} \cmidrule(lr){4-5} \cmidrule(lr){6-7}
Method & AAR $\%$ ($\uparrow$) & RMSD ($\downarrow$) & AAR $\%$ ($\uparrow$) & RMSD ($\downarrow$)  & AAR $\%$ ($\uparrow$) & RMSD ($\downarrow$) \\
\hline
LSTM & 40.98 $\pm$ \textcolor{gray}{5.20} & \textcolor{gray}{(N/A)} & 28.50 $\pm$ \textcolor{gray}{1.55} & \textcolor{gray}{(N/A)} & 15.69 $\pm$ \textcolor{gray}{0.91} & \textcolor{gray}{(N/A)}\\
C-LSTM & 40.93 $\pm$ \textcolor{gray}{5.41} & \textcolor{gray}{(N/A)} & 29.24 $\pm$ \textcolor{gray}{1.08}& \textcolor{gray}{(N/A)} & 15.48 $\pm$ \textcolor{gray}{1.17} & \textcolor{gray}{(N/A)}\\
RefineGNN & 39.40 $\pm$ \textcolor{gray}{5.56} & 3.22 $\pm$ \textcolor{gray}{0.29} & 37.06 $\pm$ \textcolor{gray}{3.09} & 3.64 $\pm$ \textcolor{gray}{0.40} & 21.13 $\pm$ \textcolor{gray}{1.59}& 6.00 $\pm$ \textcolor{gray}{0.55}  \\
C-RefineGNN & 33.19 $\pm$ \textcolor{gray}{2.99} & 3.25 $\pm$ \textcolor{gray}{0.40} & 33.53 $\pm$ \textcolor{gray}{3.23}& 3.69 $\pm$ \textcolor{gray}{0.56} & 18.88 $\pm$ \textcolor{gray}{1.37} & 6.22 $\pm$ \textcolor{gray}{0.59} \\
MEAN & 58.29 $\pm$ \textcolor{gray}{7.27} & 0.98 $\pm$ \textcolor{gray}{0.16} & 47.15 $\pm$ \textcolor{gray}{3.09} & 0.95 $\pm$ \textcolor{gray}{0.05}& 36.38 $\pm$ \textcolor{gray}{3.08} & 2.21 $\pm$ \textcolor{gray}{0.16}  \\
\abode &  \textbf{70.5} $\pm$ \textcolor{gray}{1.14} & \textbf{0.65} $\pm$ \textcolor{gray}{0.1} & \textbf{55.7} $\pm$ \textcolor{gray}{1.45} &\textbf{0.73} $\pm$ \textcolor{gray}{0.14}   & \textbf{39.8} $\pm$ \textcolor{gray}{1.17}  & \textbf{1.73} $\pm$ \textcolor{gray}{0.11}  \\
\hline
\end{tabular}}
\vskip -0.1in
\end{table*}

\subsection{Training Objective}
We optimize for the data fit of the generated states z(T) given by the differential function $f_{\psi}$. The loss consists of two components: one for the sequence and another for the structure
\begin{align}
    \calL = \calL_{\texttt{seq}} + \calL_{\texttt{structure}}
\end{align}
The sequence loss is quantified in terms of the cross-entropy between the true label $\a_{ni}^{\texttt{true}}$ and the label distribution $\a_{ni}$predicted by the model, i.e.,
\begin{align}\label{eq:entropy}
    \calL_{\texttt{seq}} =\frac{1}{N} \sum_{n=1}^N \frac{1}{M} \sum_{i=1}^{M_i} \mathrm{CE}\left(\mathbf{a}_{n i}^{\text {true }}, \mathbf{a}_{n i}\right)
\end{align}
where $n$ indexes the $N$ datapoints and $i$ indexes the $M_{i}$ residues. The structure loss is computed based on the fit to the data sample in terms of the angles and radii:
\begin{align}
    \calL_{\texttt{structure}} =\frac{1}{N} \sum_{n=1}^N \frac{1}{M}\sum_{i=1}^{M_i} \lambda\left(\calL_{\texttt{angle}}^{ni} + \calL_{\texttt{radius}}^{ni} \right).
\end{align}
For each residue angle pair $(\alpha,\gamma)$ we compute the negative log of the von-Mises likelihood
\begin{align}
   \calL_{\texttt{angle}}^{ni} = \sum_k^{\left\{\mathrm{C}_\alpha, \mathrm{C}, \mathrm{N}\right\}} \sum_{\theta \in\{\alpha, \gamma\}} \log \mathcal{M}\left(\theta_{i k}^n \mid \theta_{i k}^{n, \text { true }}, \kappa\right) 
\end{align}
where $\kappa$ is a scale parameter, and $k$ is atom index. The von Mises distribution can be interpreted as a Gaussian distribution over the domain of angles. On the other hand, the radius loss is the negative log of a Gaussian distance.
\begin{align}
   \calL_{\texttt{radius}}^{ni} = \sum_k^{\left\{\mathrm{c}_\alpha, \mathrm{C}, \mathrm{N}\right\}} \log \mathcal{N}\left(r_{i k}^n \mid r_{i k}^{n, \text { true }}, \sigma_r^2\right)
\end{align}
where $\sigma_r^2$ is the radius variance. Note that our method
predicts the sidechain spatial coordinates, also used to calculate the total loss. Here $\lambda$ is the polar loss weight, set to $\lambda=0.8$. We set $\kappa=10$, $\sigma_r^2 = 0.1$ to prefer narrow likelihoods for accurate structure prediction.

We next describe the generation step.
\subsection{Sequence and structure prediction}
Given the antibody or antigen-antibody complex, we generate an antibody sequence and the corresponding structure by solving the system of ODEs as described in section~\ref{sec:ode} for time T to obtain $\z(T ) = [\a(T ), \s(T )]$. Using the softmax operator, we transform the label features $\a(T )$ into Categorical amino acid probabilities $\textbf{p}$. We pick the most probable amino acid per node. A schematic representation is shown in Fig.~\ref{fig:work}
\section{Experiments}
\paragraph{Tasks} We benchmark \abode~on a series of challenging tasks: (i) we evaluate the model on unconditional antibody sequence and structure generation against ground truth structures in the Structural Antibody Database SAbDab~\cite{dunbar2014sabdab} section~\ref{sec:uncond}, (ii) we benchmark our method in terms of its ability to generate antigen-conditioned anti- body sequences and structures from SAbDab in section~\ref{sec:cond}, (iii) we evaluate our model on the task of designing CDR- H3 over 60 manually selected diverse complexes~\cite{rabd} in section~\ref{sec:rabd}, (iv) we extend our model to incorporate information about the constant region of the antibody in section~\ref{sec:cond_whole}, and finally, (v) we extend AbODE to de novo protein sequence design with a fixed backbone in section~\ref{sec:prot_design}.
\paragraph{Baselines} We compare \abode~ with the state-of-the-art baseline methods. On the uncontrolled generation task, we compare against sequence-only \textbf{LSTM}~\cite{saka2021antibody,akbar2022silico}, an autoregressive graph network AR- GNN~\cite{you2018graphrnn} tailored for antibodies, and an autoregressive method RefineGNN~\cite{refinegnn}, which considers the 3D geometry and co-models the sequence and the structure.

On the antigen-conditioned sequence and structure genera- tion task, we again compare against LSTM and RefineGNN. We also consider their variants C-LSTM and C-RefineGNN proposed in \citet{mean}, where they adapt the current methodology to consider the entire context of the antibody-antigen complex. We additionally consider \textbf{MEAN}~\cite{mean} which uses progressive decoding to generate CDR by encoding the external antigen context of 1D/3D information. Finally, we also compare against a physics-based simulator RosettaAD~\cite{rabd}.

\paragraph{Implementation} AbODE is implemented in PyTorch~\cite{paszke2019pytorch}. We used three layers of a Transformer Convolutional Network~\cite{shi2020masked} with embedding dimensions of $128-256-64$. Our models were trained with the Adam optimizer for 5000 epochs using batch size 300. For details, we refer the reader to Appendix~\ref{sec:implement}.
\begin{table*}[!t]
\centering
\vskip 0.1in
\caption{\textbf{Top}: Adding constant region information for unconditioned sequence and structure modeling task. \textbf{Bottom}: Adding constant region information for antigen-conditioned antibody sequence and structure modeling task.}
\label{tab:result_whole}
\centering
\resizebox{0.88\textwidth}{!}{
\begin{tabular}{lcccccc}
\hline
& \multicolumn{2}{c}{CDR-H1} & \multicolumn{2}{c}{CDR-H2}  & \multicolumn{2}{c}{CDR-H3}  \\
\cmidrule(lr){2-3} \cmidrule(lr){4-5} \cmidrule(lr){6-7}
\abode & PPL ($\downarrow$) & RMSD ($\downarrow$) & PPL ($\downarrow$) & RMSD ($\downarrow$)  & PPL ($\downarrow$) & RMSD ($\downarrow$)  \\
\hline
$-$ Constant Region &  4.25 $\pm$ \textcolor{gray}{0.46} & 0.73 $\pm$ \textcolor{gray}{0.15} & 4.32 $\pm$ \textcolor{gray}{0.32} &0.63 $\pm$ \textcolor{gray}{0.19}   & 6.35 $\pm$ \textcolor{gray}{0.29}  & 2.01 $\pm$ \textcolor{gray}{0.13}   \\
$+$ Constant Region ($z_{<i}$) &  4.31 $\pm$ \textcolor{gray}{0.31} & 0.69 $\pm$ \textcolor{gray}{0.21} & 4.17 $\pm$ \textcolor{gray}{0.29} &0.59 $\pm$ \textcolor{gray}{0.21}   & 6.41 $\pm$ \textcolor{gray}{0.37}  & 1.94 $\pm$ \textcolor{gray}{0.17}    \\
\hline
\end{tabular}}
\vskip 0.1in
\resizebox{0.88\textwidth}{!}{
\begin{tabular}{lcccccc}
\hline
& \multicolumn{2}{c}{CDR-H1} & \multicolumn{2}{c}{CDR-H2}  & \multicolumn{2}{c}{CDR-H3}  \\
\cmidrule(lr){2-3} \cmidrule(lr){4-5} \cmidrule(lr){6-7}
\abode & AAR $\%$ ($\uparrow$) & RMSD ($\downarrow$) & AAR $\%$ ($\uparrow$) & RMSD ($\downarrow$)  & AAR $\%$ ($\uparrow$) & RMSD ($\downarrow$)  \\
\hline
$-$ Constant Region &  70.5 $\pm$ \textcolor{gray}{1.14} & 0.65 $\pm$ \textcolor{gray}{0.1} & 55.7 $\pm$ \textcolor{gray}{1.45} &0.73 $\pm$ \textcolor{gray}{0.14}   &39.8 $\pm$ \textcolor{gray}{1.17}  & 1.73 $\pm$ \textcolor{gray}{0.11} \\
$+$ Constant Region ($z_{<i}$) &  71.9 $\pm$ \textcolor{gray}{1.87} & 0.71 $\pm$ \textcolor{gray}{0.23} & 56.8 $\pm$ \textcolor{gray}{1.97} &0.70 $\pm$ \textcolor{gray}{0.14}   & 36.7 $\pm$ \textcolor{gray}{1.5}  & 1.88 $\pm$ \textcolor{gray}{0.11}   \\
\hline
\end{tabular}}
\vskip -0.1in
\end{table*}
\subsection{Unconditioned Sequence and Structure Modeling} \label{sec:uncond}

\paragraph{Data} We obtained the antibody sequences and structure from Structural Antibody Database (SAbDab)~\cite{dunbar2014sabdab} and removed any incomplete or redundant complexes. We followed a similar strategy to \citet{refinegnn}, where we focus on generating heavy chain CDRs, and curated the dataset by clustering the CDR sequences via MMseq2~\cite{steinegger2017mmseqs2} with $40\%$ sequence identity. We then randomly split the clusters into training, validation, and test sets with an 8:1:1 ratio.
\paragraph{Metrics} We evaluate our method on perplexity (PPL) and root mean square deviation (RMSD) between the predicted structures and the ground truth structures on the test data. We report the results for all the CDR-H regions. We calculate the RMSD by the Kabsch algorithm~\cite{kabsch1976solution} based on $C_{\alpha}$ spatial features of the CDR residues.

\begin{figure*}[!tbp]
\begin{center}
\centerline{\includegraphics[width=0.9\textwidth]{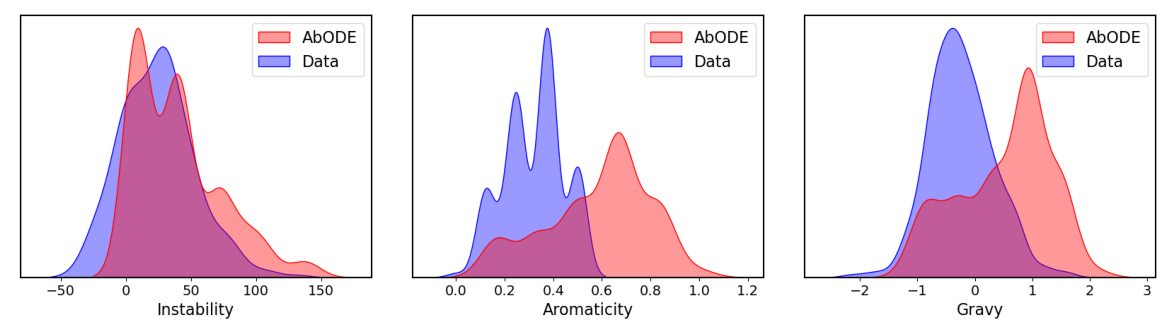}}
\caption{Functional evaluation of generated antibodies vs. data for CDR-H1 unconditional antibody sequence and structure design}
\label{fig:function_eval}
\end{center}
\end{figure*}

\paragraph{Results} The LSTM baselines do not involve structure prediction, so we only report the RMSD for the graph-based method. Table~\ref{tab:result_cond_uncond} reports the performance of \abode~ on uncontrolled generation, where \abode~ outperforms all the baselines on both metrics. Notably, \abode~ significantly reduces the PPL in all CDR regions and typically predicts a structure close to the ground truth structure. We also evaluate the biological functionality of the generated antibodies, shown in Fig.~\ref{fig:function_eval}. Specifically, we considered the following properties:
\begin{itemize}
    \item \textbf{Gravy}: The Gravy value is calculated by adding the hydropathy value for each residue and dividing it by the length of the sequence~\citep{kyte1982simple}
    \item \textbf{Instability}: The Instability index is calculated using the approach of~\citet{guruprasad1990correlation}, which predicts regional instability of dipeptides that occur more frequently in unstable proteins when compared to stable proteins.
    \item \textbf{Aromaticity}: It calculates the aromaticity value of a protein according to~\citet{lobry1994hydrophobicity}. It is simply the relative frequency of Phe+Trp+Tyr.
\end{itemize}
As our plots demonstrate, AbODE can essentially replicate the behavior of the data  in terms of instability and gravy. However, there is some discrepancy in terms of spread concerning aromaticity. 

\subsection{Antigen Conditioned Sequence and Structure Modeling} \label{sec:cond}
\paragraph{Data} We took the antigen-antibody complexes dataset from Structural Antibody Database~\cite{dunbar2014sabdab} and removed the illegal data points, renumbering them to the IMGT scheme~\cite{lefranc2003imgt}. We follow the data preparation strategy of \citet{mean,refinegnn} by splitting the dataset into training, validation, and test sets. We accomplish this by clustering the sequences via MMseq2~\cite{steinegger2017mmseqs2} with $40\%$ sequence identity. Then we split all clusters into training, validation, and test sets in the proportion 8:1:1.

\paragraph{Metrics} We employ Amino Acid Recovery (AAR) and RMSD for quantitative evaluation. AAR is defined as the overlapping rate between the predicted 1D sequences and the ground truth. RMSD is calculated via the Kabsch algorithm~\cite{kabsch1976solution} based on $C_{\alpha}$ spatial features of the CDR residues.

\paragraph{Results} Table~\ref{tab:result_cond_uncond} shows the performance of \abode~ compared to the baseline methods. \abode~ is able to perform better than other competing methods in terms of structure and sequence prediction. \abode~ is able to improve over the SOTA by directly combining the antibody context with the information about the antigen via the attention network, thereby demonstrating the benefits of joint modeling. As a result, \abode~ able to learn the underlying distribution of the complexes effectively.
\begin{table*}[!t]
\centering
\vskip 0.1in
\caption{ Perplexity (PPL) and Amino Acid Recovery (AAR) for different methods on fixed backbone sequence design task. Baselines are from~\citet{shi2023protein}.}
\label{tab:result_backbone}
\centering
\resizebox{0.89\textwidth}{!}{
\begin{tabular}{lcccccc}
\hline
& \multicolumn{3}{c}{PPL ($\downarrow$)} & \multicolumn{3}{c}{AAR $\%$ ($\uparrow$) }   \\
\cmidrule(lr){2-4} \cmidrule(lr){5-7}
Method & Short & Single-chain & All & Short  & Single-chain & All   \\
\hline
GVP-Transformer & $8.94$& $8.67$ & $6.70$ & $27.3$ & $28.3$ & $36.5$ \\
Structured GNN & $8.31$ & $8.88$ & $6.55$ & $28.4$ & $28.1$ & $37.3$ \\
GVP-GNN & $7.10$ & $7.44$ & $5.29$ & $32.1$ & $32.0$ & $40.2$  \\
ProtSeed & $7.32$ & $7.38$ & $5.60$ & $34.8$ & $34.1$ & $43.8$  \\
\abode &  $7.19$ $\pm$ \textcolor{gray}{$0.34$} & $7.33$ $\pm$ \textcolor{gray}{$0.25$} & $5.85$ $\pm$ \textcolor{gray}{$0.45$} &$34.4$ $\pm$ \textcolor{gray}{$1.7$}   & $34.7$ $\pm$ \textcolor{gray}{$1.2$}  & $42.7$ $\pm$ \textcolor{gray}{$1.9$}    \\
\hline
\end{tabular}}
\vskip -0.1in
\end{table*}
\subsection{Antigen-Binding CDR-H3 Design}\label{sec:rabd}
In order to further evaluate our model, we designed CDR- H3 that binds to a given antigen. We used AAR and RMSD as our scoring metrics. We included RosettaAD~\cite{rabd}, a conventional physics-based baseline for comparison. We benchmark our method on 60 diverse complexes selected by~\cite{rabd}.

Note, however, that the training is still conducted on the SAbDab dataset as described in section~\ref{sec:cond}, where we eliminate the antibodies that overlap with those in RAbD to avoid any data leakage.

\paragraph{Results} The performance of \abode~ , and its comparison with the baselines, is reported in Table~\ref{tab:result_rabd}. \abode~ can improve upon the best-performing baseline MEAN while significantly outperforming all the other baselines in terms of both the AAR and the RMSD. In particular, the higher Amino acid recovery rate (AAR) of \abode~ relative to the other methods demonstrates the ability of the proposed method to learn the underlying distribution of residuals for sequence design.

\subsection{Conditional Generation given Framework Region}\label{sec:cond_whole}
We next extend the proposed method by incorporating the sequence and structural information besides the CDR regions (i.e., constant region). We encode the sequence and structure information of the residues before CDR-H1, H2, and H3. Specifically, we define a k-nearest neighbor graph over the spatial domain for residues and use the sequence $\z_{< i}$, where $i$ is the location of the first CDR-H1 (or H2/H3 as the case maybe), top obtain an encoding 
\begin{align*}
    h_{<i} &= \phi^{enc}(\z_{<i},\z_{\mathcal{N}_{<i}},\{e_{ij}\}_{j \in \mathcal{N}_{<i}}) \\
    h &= \texttt{Agg}(h_{<i})
\end{align*}
where $\z_{<i}  = [\a_{<i},\s_{<i}]$, $\mathcal{N}_{<i}$ denotes the neighbours of the
residues, and $e_{ij}$ are the edge features. We parameterize $\phi^{enc}$ as a 2-layer Transformer Convolutional Network~\cite{shi2020masked}, setting the encoding dimension to 16. The encoded features $h_{<i}$ are then aggregated to provide a single summarized representation $h$ per antibody, which is then used in dynamics
\begin{align*}
    \dz_{i}(t) = \frac{\partial \z_{i}(t)}{\partial t} = f_{\psi}(t,\z_{i}(t),\z_{N(i)}(t),\{\mathbf{e}_{ij}(t)\}_{j},h)
\end{align*}
Consequently, in this case, our method has access to extra information from the rest of the antibody sequence, leading to more nuanced dynamics. Further details are provided in Appendix~\ref{sec:whole_cond}.

\paragraph{Results} We evaluate this variant of our method on both uncontrolled antibody sequence-structure design and antigen-conditioned antibody sequence structure co-design, as described in section~\ref{sec:uncond} and \ref{sec:cond}. The performance of \abode~ is reported in Table~\ref{tab:result_whole}. We observe that including the constant region increases performance for some CDR regions.

\begin{table}[!hbt]
\centering
\vskip 0.1in
\caption{ Results on RAbD benchmark. We report Amino acid recovery (AAR) and RMSD for CDR-H3 design. Baselines are from~\citet{mean}.}
\label{tab:result_rabd}
\centering
\resizebox{0.4\textwidth}{!}{
\begin{tabular}{lcc}
\hline
Method & AAR $\%$ ($\uparrow$) & RMSD ($\downarrow$)   \\
\hline
RosettaAD & 22.50& 5.52  \\
LSTM & 22.36 & \textcolor{gray}{(N/A)} \\
C-LSTM & 22.18 & \textcolor{gray}{(N/A)}  \\
RefineGNN & 29.79 & 7.55   \\
C-RefineGNN & 28.90 & 7.21   \\
MEAN & 36.77 & 1.81   \\
\abode &  \textbf{39.95} $\pm$ \textcolor{gray}{1.3} & \textbf{1.54} $\pm$ \textcolor{gray}{0.24}    \\
\hline
\end{tabular}}
\vskip -0.1in
\end{table}

\begin{table*}[!hbt]
    \caption{ Amino acid recovery (AAR) and root mean square deviation (RMSD) for masking a certain part of antigen in antigen-conditioned antibody sequence and structure generation.}
    \label{tab:result_mask}
    \vskip 0.1in
    \begin{center}
    \resizebox{0.8\textwidth}{!}{
    \begin{tabular}{lcccccc}
\toprule
& \multicolumn{2}{c}{CDR-H1} & \multicolumn{2}{c}{CDR-H2}  & \multicolumn{2}{c}{CDR-H3}  \\
\cmidrule(lr){2-3} \cmidrule(lr){4-5} \cmidrule(lr){6-7}
\abode & AAR \% ($\uparrow$) & RMSD ($\downarrow$)  & AAR \% ($\uparrow$) & RMSD ($\downarrow$) & AAR \% ($\uparrow$) & RMSD ($\downarrow$)  \\
\midrule
Mask = $10\%$ &  63.7  & 0.87 & 49.7  & 0.88 & 33.1 & 1.99  \\ 
Mask = $0\%$ &  70.5 & 0.65 & 55.7 & 0.73 & 39.8 & 1.73   \\
\bottomrule
\end{tabular}}
    \end{center}
    \vskip -0.1in
\end{table*}  

\subsection{Fixed Backbone Sequence Design}\label{sec:prot_design}
We finally extend the evaluation of our method to design protein sequences that can fold into a given backbone structure. This task is commonly known as the fixed backbone structure design.

We utilized the protein dihedral angles and other spatial features described in Eq~\ref{eq:edge_feat} and \citet{jing2020learning}. These features can be derived solely from backbone coordinates \cite{ingraham2019generative},  as the protein structures are fixed from the beginning. We use the CATH 4.2 dataset curated by \citet{ingraham2019generative} and followed the same experimental setting as used in previous works for a fair comparison. We compare \abode~ with state-of-the-art baselines for fixed backbone design, including Structured GNN \citep{ingraham2019}, GVP-GNN \citep{jing2020learning}, GVP-Transformer \citep{hsu2022learning} and ProtSeed \citep{shi2023protein}. We evaluate the per- formance of all methods using PPL and AAR as introduced in previous sections. Additional details can be found in \ref{sec:backbone_design}.

\paragraph{Results} Results The comparison of \abode~ with other baselines is shown in Table 4. We note that \abode~ is able to perform comparably to other methods when the evaluation is performed on test splits in CATH 4.2 test set. These include the short chains that have at most 100 residues and the single-chain protein sequences. Our results establish the promise of \abode~ as a protein sequence design method (conditioned on desired backbone structures), and suggest that \abode~ may be generalizable to related tasks beyond antibody design.
\section{Ablation Studies}
\subsection{Masked-Antigen Conditioned Sequence and Structure Modeling}
We evaluated the performance of our method when data was missing. We investigate this scenario by masking $10\%$ amino acids of the antigen with the minimum number of amino acids being masked 1 (note that masking 10\% becomes especially critical when the antigen is a peptide with only 5-9 amino acids) for antigen-conditioned antibody sequence and structure generation. Table~\ref{tab:result_mask} shows the empirical results of the proposed method (\abode) on antigen-conditioned antibody sequence and structure generation  as described in section~\ref{sec:cond}. Compared to the original, unmasked setting (in Table~\ref{tab:result_cond_uncond}), we observe some dip in the performance compared to the original setting, as expected.
\begin{table}[!hbt]
\centering
\vskip 0.1in
\caption{Hyperparameter effect of the number of time steps for solving the ODE for CDR-H1 data }
\label{tab:result_time}
\centering
\resizebox{0.3\textwidth}{!}{
\begin{tabular}{lcc}
\hline
\abode &  PPL ($\downarrow$) & RMSD ($\downarrow$)  \\
\midrule
$t=10$ & 7.38  & 1.44  \\ 
$t=50$ & 7.18  & 1.87  \\
$t=200$ & 5.18  & 1.01  \\
\hline
\end{tabular}}
\end{table}
\subsection{Time hyperparameter for ODE}
We also evaluated the effect of different choices of time steps $t$ to solve our ODE system. Table~\ref{tab:result_time} demonstrates the effect of change in the number of time steps on the downstream performance for CDR-H1 data on Antigen Conditioned Sequence and Structure Modeling. We note that increasing the number of timesteps for solving the ODE increases performance and that training with fewer time steps leads to unstable training.
\section{Conclusion}
We introduced a new generative model \abode, which models the antibody-antigen complex as a joint graph and performs information propagation using a graph PDE that reduces to a system of coupled residue-specific ODEs. \abode~can accurately co-model the sequence and structure of the antigen-antibody complex. In particular, the model can generate a binding antibody sequence and structure with state-of-the-art accuracy for a given antigen.
\section*{Acknowledgements}
The calculations were performed using resources made available by the Aalto University Science-IT project. This work has been supported by the Academy of Finland under the {\em HEALED} project (grant 13342077).

\bibliography{main}

\begin{thebibliography}{58}
\providecommand{\natexlab}[1]{#1}
\providecommand{\url}[1]{\texttt{#1}}
\expandafter\ifx\csname urlstyle\endcsname\relax
  \providecommand{\doi}[1]{doi: #1}\else
  \providecommand{\doi}{doi: \begingroup \urlstyle{rm}\Url}\fi

\bibitem[Adolf-Bryfogle et~al.(2018)Adolf-Bryfogle, Kalyuzhniy, Kubitz,
  Weitzner, Hu, Adachi, Schief, and Dunbrack~Jr]{rabd}
Adolf-Bryfogle, J., Kalyuzhniy, O., Kubitz, M., Weitzner, B.~D., Hu, X.,
  Adachi, Y., Schief, W.~R., and Dunbrack~Jr, R.~L.
\newblock Rosettaantibodydesign (rabd): A general framework for computational
  antibody design.
\newblock \emph{PLoS computational biology}, 14\penalty0 (4):\penalty0
  e1006112, 2018.

\bibitem[Akbar et~al.(2022)Akbar, Robert, Weber, Widrich, Frank, Pavlovi{\'c},
  Scheffer, Chernigovskaya, Snapkov, Slabodkin, et~al.]{akbar2022silico}
Akbar, R., Robert, P.~A., Weber, C.~R., Widrich, M., Frank, R., Pavlovi{\'c},
  M., Scheffer, L., Chernigovskaya, M., Snapkov, I., Slabodkin, A., et~al.
\newblock In silico proof of principle of machine learning-based antibody
  design at unconstrained scale.
\newblock In \emph{MAbs}, volume~14, pp.\  2031482. Taylor \& Francis, 2022.

\bibitem[Alley et~al.(2019)Alley, Khimulya, Biswas, AlQuraishi, and
  Church]{prot2}
Alley, E.~C., Khimulya, G., Biswas, S., AlQuraishi, M., and Church, G.~M.
\newblock Unified rational protein engineering with sequence-based deep
  representation learning.
\newblock \emph{Nature methods}, 16\penalty0 (12):\penalty0 1315--1322, 2019.

\bibitem[Anand \& Huang(2018)Anand and Huang]{anand2018}
Anand, N. and Huang, P.
\newblock Generative modeling for protein structures.
\newblock \emph{Advances in neural information processing systems}, 31, 2018.

\bibitem[Avelar et~al.(2019)Avelar, Tavares, Gori, and
  Lamb]{avelar2019discrete}
Avelar, P.~H., Tavares, A.~R., Gori, M., and Lamb, L.~C.
\newblock Discrete and continuous deep residual learning over graphs.
\newblock \emph{arXiv preprint arXiv:1911.09554}, 2019.

\bibitem[Baek et~al.(2021)Baek, DiMaio, Anishchenko, Dauparas, Ovchinnikov,
  Lee, Wang, Cong, Kinch, Schaeffer, et~al.]{pf3}
Baek, M., DiMaio, F., Anishchenko, I., Dauparas, J., Ovchinnikov, S., Lee,
  G.~R., Wang, J., Cong, Q., Kinch, L.~N., Schaeffer, R.~D., et~al.
\newblock Accurate prediction of protein structures and interactions using a
  three-track neural network.
\newblock \emph{Science}, 373\penalty0 (6557):\penalty0 871--876, 2021.

\bibitem[Cao et~al.(2021)Cao, Das, Chenthamarakshan, Chen, Melnyk, and
  Shen]{prot5}
Cao, Y., Das, P., Chenthamarakshan, V., Chen, P.-Y., Melnyk, I., and Shen, Y.
\newblock Fold2seq: A joint sequence (1d)-fold (3d) embedding-based generative
  model for protein design.
\newblock In \emph{International Conference on Machine Learning}, pp.\
  1261--1271. PMLR, 2021.

\bibitem[Chamberlain et~al.(2021)Chamberlain, Rowbottom, Gorinova, Bronstein,
  Webb, and Rossi]{chamberlain2021grand}
Chamberlain, B., Rowbottom, J., Gorinova, M.~I., Bronstein, M., Webb, S., and
  Rossi, E.
\newblock Grand: Graph neural diffusion.
\newblock In \emph{International Conference on Machine Learning}, pp.\
  1407--1418. PMLR, 2021.

\bibitem[Chen et~al.(2018)Chen, Rubanova, Bettencourt, and
  Duvenaud]{chen2018neural}
Chen, R.~T., Rubanova, Y., Bettencourt, J., and Duvenaud, D.~K.
\newblock Neural ordinary differential equations.
\newblock \emph{Advances in neural information processing systems}, 31, 2018.

\bibitem[Corso et~al.(2023)Corso, Stärk, Jing, Barzilay, and
  Jaakkola]{diffdock}
Corso, G., Stärk, H., Jing, B., Barzilay, R., and Jaakkola, T.
\newblock Diffdock: Diffusion steps, twists, and turns for molecular docking,
  2023.

\bibitem[Dauparas et~al.(2022)Dauparas, Anishchenko, Bennett, Bai, Ragotte,
  Milles, Wicky, Courbet, de~Haas, Bethel, et~al.]{protmpnn}
Dauparas, J., Anishchenko, I., Bennett, N., Bai, H., Ragotte, R.~J., Milles,
  L.~F., Wicky, B.~I., Courbet, A., de~Haas, R.~J., Bethel, N., et~al.
\newblock Robust deep learning--based protein sequence design using
  proteinmpnn.
\newblock \emph{Science}, 378\penalty0 (6615):\penalty0 49--56, 2022.

\bibitem[Dunbar et~al.(2014)Dunbar, Krawczyk, Leem, Baker, Fuchs, Georges, Shi,
  and Deane]{dunbar2014sabdab}
Dunbar, J., Krawczyk, K., Leem, J., Baker, T., Fuchs, A., Georges, G., Shi, J.,
  and Deane, C.~M.
\newblock Sabdab: the structural antibody database.
\newblock \emph{Nucleic acids research}, 42\penalty0 (D1):\penalty0
  D1140--D1146, 2014.

\bibitem[Eliasof et~al.(2021)Eliasof, Haber, and Treister]{eliasof2021pde}
Eliasof, M., Haber, E., and Treister, E.
\newblock Pde-gcn: Novel architectures for graph neural networks motivated by
  partial differential equations.
\newblock \emph{Advances in Neural Information Processing Systems}, 34, 2021.

\bibitem[Fischman \& Ofran(2018)Fischman and Ofran]{fischman2018computational}
Fischman, S. and Ofran, Y.
\newblock Computational design of antibodies.
\newblock \emph{Current opinion in structural biology}, 51:\penalty0 156--162,
  2018.

\bibitem[Ganea et~al.(2021)Ganea, Huang, Bunne, Bian, Barzilay, Jaakkola, and
  Krause]{equidock}
Ganea, O.-E., Huang, X., Bunne, C., Bian, Y., Barzilay, R., Jaakkola, T., and
  Krause, A.
\newblock Independent se (3)-equivariant models for end-to-end rigid protein
  docking.
\newblock \emph{arXiv preprint arXiv:2111.07786}, 2021.

\bibitem[Gao et~al.(2022)Gao, Tan, and Li]{alphadesign}
Gao, Z., Tan, C., and Li, S.~Z.
\newblock Alphadesign: A graph protein design method and benchmark on
  alphafolddb, 2022.

\bibitem[Grathwohl et~al.(2018)Grathwohl, Chen, Bettencourt, Sutskever, and
  Duvenaud]{grathwohl2018ffjord}
Grathwohl, W., Chen, R.~T., Bettencourt, J., Sutskever, I., and Duvenaud, D.
\newblock Ffjord: Free-form continuous dynamics for scalable reversible
  generative models.
\newblock \emph{arXiv preprint arXiv:1810.01367}, 2018.

\bibitem[Graves et~al.(2020)Graves, Byerly, Priego, Makkapati, Parish,
  Medellin, and Berrondo]{ant4}
Graves, J., Byerly, J., Priego, E., Makkapati, N., Parish, S.~V., Medellin, B.,
  and Berrondo, M.
\newblock A review of deep learning methods for antibodies.
\newblock \emph{Antibodies}, 9\penalty0 (2):\penalty0 12, 2020.

\bibitem[Guruprasad et~al.(1990)Guruprasad, Reddy, and
  Pandit]{guruprasad1990correlation}
Guruprasad, K., Reddy, B.~B., and Pandit, M.~W.
\newblock Correlation between stability of a protein and its dipeptide
  composition: a novel approach for predicting in vivo stability of a protein
  from its primary sequence.
\newblock \emph{Protein Engineering, Design and Selection}, 4\penalty0
  (2):\penalty0 155--161, 1990.

\bibitem[Hsu et~al.(2022)Hsu, Verkuil, Liu, Lin, Hie, Sercu, Lerer, and
  Rives]{hsu2022learning}
Hsu, C., Verkuil, R., Liu, J., Lin, Z., Hie, B., Sercu, T., Lerer, A., and
  Rives, A.
\newblock Learning inverse folding from millions of predicted structures.
\newblock In \emph{International Conference on Machine Learning}, pp.\
  8946--8970. PMLR, 2022.

\bibitem[Iakovlev et~al.(2020)Iakovlev, Heinonen, and
  L{\"a}hdesm{\"a}ki]{iakovlev2020learning}
Iakovlev, V., Heinonen, M., and L{\"a}hdesm{\"a}ki, H.
\newblock Learning continuous-time pdes from sparse data with graph neural
  networks.
\newblock \emph{arXiv preprint arXiv:2006.08956}, 2020.

\bibitem[Ingraham et~al.(2019{\natexlab{a}})Ingraham, Garg, Barzilay, and
  Jaakkola]{ingraham2019}
Ingraham, J., Garg, V., Barzilay, R., and Jaakkola, T.
\newblock Generative models for graph-based protein design.
\newblock In Wallach, H., Larochelle, H., Beygelzimer, A., d\textquotesingle
  Alch\'{e}-Buc, F., Fox, E., and Garnett, R. (eds.), \emph{Advances in Neural
  Information Processing Systems}, volume~32. Curran Associates, Inc.,
  2019{\natexlab{a}}.
\newblock URL
  \url{https://proceedings.neurips.cc/paper/2019/file/f3a4ff4839c56a5f460c88cce3666a2b-Paper.pdf}.

\bibitem[Ingraham et~al.(2019{\natexlab{b}})Ingraham, Garg, Barzilay, and
  Jaakkola]{ingraham2019generative}
Ingraham, J., Garg, V., Barzilay, R., and Jaakkola, T.
\newblock Generative models for graph-based protein design.
\newblock \emph{Advances in neural information processing systems}, 32,
  2019{\natexlab{b}}.

\bibitem[Ingraham et~al.(2019{\natexlab{c}})Ingraham, Riesselman, Sander, and
  Marks]{pf1}
Ingraham, J., Riesselman, A., Sander, C., and Marks, D.
\newblock Learning protein structure with a differentiable simulator.
\newblock In \emph{International Conference on Learning Representations},
  2019{\natexlab{c}}.

\bibitem[Ingraham et~al.(2019{\natexlab{d}})Ingraham, Riesselman, Sander, and
  Marks]{pf2}
Ingraham, J., Riesselman, A., Sander, C., and Marks, D.
\newblock Learning protein structure with a differentiable simulator.
\newblock In \emph{International Conference on Learning Representations},
  2019{\natexlab{d}}.

\bibitem[Ingraham et~al.(2022)Ingraham, Baranov, Costello, Frappier, Ismail,
  Tie, Wang, Xue, Obermeyer, Beam, and Grigoryan]{pf5}
Ingraham, J., Baranov, M., Costello, Z., Frappier, V., Ismail, A., Tie, S.,
  Wang, W., Xue, V., Obermeyer, F., Beam, A., and Grigoryan, G.
\newblock Illuminating protein space with a programmable generative model.
\newblock \emph{bioRxiv}, 2022.
\newblock \doi{10.1101/2022.12.01.518682}.
\newblock URL
  \url{https://www.biorxiv.org/content/early/2022/12/02/2022.12.01.518682}.

\bibitem[Jin et~al.(2022{\natexlab{a}})Jin, Barzilay, and
  Jaakkola]{pmlr-v162-jin22a}
Jin, W., Barzilay, D., and Jaakkola, T.
\newblock Antibody-antigen docking and design via hierarchical structure
  refinement.
\newblock In Chaudhuri, K., Jegelka, S., Song, L., Szepesvari, C., Niu, G., and
  Sabato, S. (eds.), \emph{Proceedings of the 39th International Conference on
  Machine Learning}, volume 162 of \emph{Proceedings of Machine Learning
  Research}, pp.\  10217--10227. PMLR, 17--23 Jul 2022{\natexlab{a}}.
\newblock URL \url{https://proceedings.mlr.press/v162/jin22a.html}.

\bibitem[Jin et~al.(2022{\natexlab{b}})Jin, Wohlwend, Barzilay, and
  Jaakkola]{refinegnn}
Jin, W., Wohlwend, J., Barzilay, R., and Jaakkola, T.
\newblock Iterative refinement graph neural network for antibody
  sequence-structure co-design, 2022{\natexlab{b}}.

\bibitem[Jing et~al.(2020)Jing, Eismann, Suriana, Townshend, and
  Dror]{jing2020learning}
Jing, B., Eismann, S., Suriana, P., Townshend, R.~J., and Dror, R.
\newblock Learning from protein structure with geometric vector perceptrons.
\newblock \emph{arXiv preprint arXiv:2009.01411}, 2020.

\bibitem[Jumper et~al.(2021)Jumper, Evans, Pritzel, Green, Figurnov,
  et~al.]{pf4}
Jumper, J., Evans, R., Pritzel, A., Green, T., Figurnov, M., et~al.
\newblock Highly accurate protein structure prediction with alphafold.
\newblock \emph{Nature}, 596\penalty0 (7873):\penalty0 583--589, 2021.

\bibitem[Kabsch(1976)]{kabsch1976solution}
Kabsch, W.
\newblock A solution for the best rotation to relate two sets of vectors.
\newblock \emph{Acta Crystallographica Section A: Crystal Physics, Diffraction,
  Theoretical and General Crystallography}, 32\penalty0 (5):\penalty0 922--923,
  1976.

\bibitem[Karimi et~al.(2020)Karimi, Zhu, Cao, and Shen]{prot4}
Karimi, M., Zhu, S., Cao, Y., and Shen, Y.
\newblock De novo protein design for novel folds using guided conditional
  wasserstein generative adversarial networks.
\newblock \emph{Journal of chemical information and modeling}, 60\penalty0
  (12):\penalty0 5667--5681, 2020.

\bibitem[Kong et~al.(2023)Kong, Huang, and Liu]{mean}
Kong, X., Huang, W., and Liu, Y.
\newblock Conditional antibody design as 3d equivariant graph translation,
  2023.

\bibitem[Kuroda et~al.(2012)Kuroda, Shirai, Jacobson, and Nakamura]{kuroda}
Kuroda, D., Shirai, H., Jacobson, M.~P., and Nakamura, H.
\newblock {Computer-aided antibody design}.
\newblock \emph{Protein Engineering, Design and Selection}, 25\penalty0
  (10):\penalty0 507--522, 06 2012.
\newblock ISSN 1741-0126.
\newblock \doi{10.1093/protein/gzs024}.
\newblock URL \url{https://doi.org/10.1093/protein/gzs024}.

\bibitem[Kyte \& Doolittle(1982)Kyte and Doolittle]{kyte1982simple}
Kyte, J. and Doolittle, R.~F.
\newblock A simple method for displaying the hydropathic character of a
  protein.
\newblock \emph{Journal of molecular biology}, 157\penalty0 (1):\penalty0
  105--132, 1982.

\bibitem[Lapidoth et~al.(2015)Lapidoth, Baran, Pszolla, Norn, Alon, Tyka, and
  Fleishman]{ant3}
Lapidoth, G.~D., Baran, D., Pszolla, G.~M., Norn, C., Alon, A., Tyka, M.~D.,
  and Fleishman, S.~J.
\newblock Abdesign: A n algorithm for combinatorial backbone design guided by
  natural conformations and sequences.
\newblock \emph{Proteins: Structure, Function, and Bioinformatics}, 83\penalty0
  (8):\penalty0 1385--1406, 2015.

\bibitem[Lefranc et~al.(2003)Lefranc, Pommi{\'e}, Ruiz, Giudicelli, Foulquier,
  Truong, Thouvenin-Contet, and Lefranc]{lefranc2003imgt}
Lefranc, M.-P., Pommi{\'e}, C., Ruiz, M., Giudicelli, V., Foulquier, E.,
  Truong, L., Thouvenin-Contet, V., and Lefranc, G.
\newblock Imgt unique numbering for immunoglobulin and t cell receptor variable
  domains and ig superfamily v-like domains.
\newblock \emph{Developmental \& Comparative Immunology}, 27\penalty0
  (1):\penalty0 55--77, 2003.

\bibitem[Li et~al.(2014)Li, Pantazes, and Maranas]{ant2}
Li, T., Pantazes, R.~J., and Maranas, C.~D.
\newblock Optmaven--a new framework for the de novo design of antibody variable
  region models targeting specific antigen epitopes.
\newblock \emph{PloS one}, 9\penalty0 (8):\penalty0 e105954, 2014.

\bibitem[Li et~al.(2019)Li, Gu, Dullien, Vinyals, and Kohli]{gmm}
Li, Y., Gu, C., Dullien, T., Vinyals, O., and Kohli, P.
\newblock Graph matching networks for learning the similarity of graph
  structured objects.
\newblock In \emph{International conference on machine learning}, pp.\
  3835--3845. PMLR, 2019.

\bibitem[Lobry \& Gautier(1994)Lobry and Gautier]{lobry1994hydrophobicity}
Lobry, J. and Gautier, C.
\newblock Hydrophobicity, expressivity and aromaticity are the major trends of
  amino-acid usage in 999 escherichia coli chromosome-encoded genes.
\newblock \emph{Nucleic acids research}, 22\penalty0 (15):\penalty0 3174--3180,
  1994.

\bibitem[O'Connell et~al.(2018)O'Connell, Li, Hanson, Heffernan, Lyons,
  Paliwal, Dehzangi, Yang, and Zhou]{prot1}
O'Connell, J., Li, Z., Hanson, J., Heffernan, R., Lyons, J., Paliwal, K.,
  Dehzangi, A., Yang, Y., and Zhou, Y.
\newblock Spin2: Predicting sequence profiles from protein structures using
  deep neural networks.
\newblock \emph{Proteins: Structure, Function, and Bioinformatics}, 86\penalty0
  (6):\penalty0 629--633, 2018.

\bibitem[Pantazes \& Maranas(2010)Pantazes and Maranas]{ant1}
Pantazes, R. and Maranas, C.~D.
\newblock Optcdr: a general computational method for the design of antibody
  complementarity determining regions for targeted epitope binding.
\newblock \emph{Protein Engineering, Design \& Selection}, 23\penalty0
  (11):\penalty0 849--858, 2010.

\bibitem[Paszke et~al.(2019)Paszke, Gross, Massa, Lerer, Bradbury, Chanan,
  Killeen, Lin, Gimelshein, Antiga, et~al.]{paszke2019pytorch}
Paszke, A., Gross, S., Massa, F., Lerer, A., Bradbury, J., Chanan, G., Killeen,
  T., Lin, Z., Gimelshein, N., Antiga, L., et~al.
\newblock Pytorch: An imperative style, high-performance deep learning library.
\newblock \emph{Advances in neural information processing systems}, 32, 2019.

\bibitem[Pinto et~al.(2020)Pinto, Park, Beltramello, Walls, Tortorici, Bianchi,
  Jaconi, Culap, Zatta, De~Marco, et~al.]{pinto2020}
Pinto, D., Park, Y.-J., Beltramello, M., Walls, A.~C., Tortorici, M.~A.,
  Bianchi, S., Jaconi, S., Culap, K., Zatta, F., De~Marco, A., et~al.
\newblock Cross-neutralization of sars-cov-2 by a human monoclonal sars-cov
  antibody.
\newblock \emph{Nature}, 583\penalty0 (7815):\penalty0 290--295, 2020.

\bibitem[Raybould et~al.(2019)Raybould, Marks, Krawczyk, Taddese, Nowak, Lewis,
  Bujotzek, Shi, and Deane]{Raybould}
Raybould, M.~I., Marks, C., Krawczyk, K., Taddese, B., Nowak, J., Lewis, A.~P.,
  Bujotzek, A., Shi, J., and Deane, C.~M.
\newblock Five computational developability guidelines for therapeutic antibody
  profiling.
\newblock \emph{Proceedings of the National Academy of Sciences}, 116\penalty0
  (10):\penalty0 4025--4030, 2019.

\bibitem[Saka et~al.(2021{\natexlab{a}})Saka, Kakuzaki, Metsugi, Kashiwagi,
  Yoshida, Wada, Tsunoda, and Teramoto]{antgen2}
Saka, K., Kakuzaki, T., Metsugi, S., Kashiwagi, D., Yoshida, K., Wada, M.,
  Tsunoda, H., and Teramoto, R.
\newblock Antibody design using lstm based deep generative model from phage
  display library for affinity maturation.
\newblock \emph{Scientific reports}, 11\penalty0 (1):\penalty0 1--13,
  2021{\natexlab{a}}.

\bibitem[Saka et~al.(2021{\natexlab{b}})Saka, Kakuzaki, Metsugi, Kashiwagi,
  Yoshida, Wada, Tsunoda, and Teramoto]{saka2021antibody}
Saka, K., Kakuzaki, T., Metsugi, S., Kashiwagi, D., Yoshida, K., Wada, M.,
  Tsunoda, H., and Teramoto, R.
\newblock Antibody design using lstm based deep generative model from phage
  display library for affinity maturation.
\newblock \emph{Scientific reports}, 11\penalty0 (1):\penalty0 1--13,
  2021{\natexlab{b}}.

\bibitem[Satorras et~al.(2021)Satorras, Hoogeboom, and Welling]{egnn}
Satorras, V.~G., Hoogeboom, E., and Welling, M.
\newblock E(n) equivariant graph neural networks, 2021.
\newblock URL \url{https://arxiv.org/abs/2102.09844}.

\bibitem[Shi et~al.(2023)Shi, Wang, Lu, Zhong, and Tang]{shi2023protein}
Shi, C., Wang, C., Lu, J., Zhong, B., and Tang, J.
\newblock Protein sequence and structure co-design with equivariant
  translation, 2023.

\bibitem[Shi et~al.(2020)Shi, Huang, Feng, Zhong, Wang, and Sun]{shi2020masked}
Shi, Y., Huang, Z., Feng, S., Zhong, H., Wang, W., and Sun, Y.
\newblock Masked label prediction: Unified message passing model for
  semi-supervised classification.
\newblock \emph{arXiv preprint arXiv:2009.03509}, 2020.

\bibitem[Shin et~al.(2021{\natexlab{a}})Shin, Riesselman, Kollasch, McMahon,
  Simon, Sander, Manglik, Kruse, and Marks]{antgen1}
Shin, J.-E., Riesselman, A.~J., Kollasch, A.~W., McMahon, C., Simon, E.,
  Sander, C., Manglik, A., Kruse, A.~C., and Marks, D.~S.
\newblock Protein design and variant prediction using autoregressive generative
  models.
\newblock \emph{Nature communications}, 12\penalty0 (1):\penalty0 2403,
  2021{\natexlab{a}}.

\bibitem[Shin et~al.(2021{\natexlab{b}})Shin, Riesselman, Kollasch, McMahon,
  Simon, Sander, Manglik, Kruse, and Marks]{shin2021protein}
Shin, J.-E., Riesselman, A.~J., Kollasch, A.~W., McMahon, C., Simon, E.,
  Sander, C., Manglik, A., Kruse, A.~C., and Marks, D.~S.
\newblock Protein design and variant prediction using autoregressive generative
  models.
\newblock \emph{Nature communications}, 12\penalty0 (1):\penalty0 2403,
  2021{\natexlab{b}}.

\bibitem[St{\"a}rk et~al.(2022)St{\"a}rk, Ganea, Pattanaik, Barzilay, and
  Jaakkola]{equibind}
St{\"a}rk, H., Ganea, O., Pattanaik, L., Barzilay, R., and Jaakkola, T.
\newblock Equibind: Geometric deep learning for drug binding structure
  prediction.
\newblock In \emph{International Conference on Machine Learning}, pp.\
  20503--20521. PMLR, 2022.

\bibitem[Steinegger \& S{\"o}ding(2017)Steinegger and
  S{\"o}ding]{steinegger2017mmseqs2}
Steinegger, M. and S{\"o}ding, J.
\newblock Mmseqs2 enables sensitive protein sequence searching for the analysis
  of massive data sets.
\newblock \emph{Nature biotechnology}, 35\penalty0 (11):\penalty0 1026--1028,
  2017.

\bibitem[Strokach et~al.(2019)Strokach, Becerra, Corbi-Verge, Perez-Riba, and
  Kim]{prot3}
Strokach, A., Becerra, D., Corbi-Verge, C., Perez-Riba, A., and Kim, P.~M.
\newblock Fast and flexible design of novel proteins using graph neural
  networks.
\newblock \emph{BioRxiv}, pp.\  868935, 2019.

\bibitem[Verma et~al.(2022)Verma, Kaski, Heinonen, and Garg]{modflow}
Verma, Y., Kaski, S., Heinonen, M., and Garg, V.
\newblock Modular flows: Differential molecular generation.
\newblock \emph{arXiv preprint arXiv:2210.06032}, 2022.

\bibitem[Wu et~al.(2022)Wu, Yang, van~den Berg, Zou, Lu, and Amini]{folding}
Wu, K.~E., Yang, K.~K., van~den Berg, R., Zou, J.~Y., Lu, A.~X., and Amini,
  A.~P.
\newblock Protein structure generation via folding diffusion, 2022.

\bibitem[You et~al.(2018)You, Ying, Ren, Hamilton, and
  Leskovec]{you2018graphrnn}
You, J., Ying, R., Ren, X., Hamilton, W., and Leskovec, J.
\newblock Graphrnn: Generating realistic graphs with deep auto-regressive
  models.
\newblock In \emph{International conference on machine learning}, pp.\
  5708--5717. PMLR, 2018.

\end{thebibliography}
\bibliographystyle{icml2023}

\newpage
\appendix
\onecolumn
\section{Appendix}
\subsection{Orientation Matrix}
\label{sec:orient}
Orientation matrix\cite{ingraham2019generative} defines invariant and locally informative features, using a local coordinate system at each residue $i$, in terms of the backbone geometry. It is formally defined as,
\begin{align}
    \mathcal{O}_{i} &= [\u_{i},\mathbf{n}_{i},\mathbf{b}_{i} \times \mathbf{n}_{i}] \\
   \u_{i} = \frac{\x_{i} - \x_{i-1}}{||\x_{i} - \x_{i-1}||},~\mathbf{b}_{i}& = \frac{\u_{i} - \u_{i+1}}{||\u_{i} - \u_{i+1}||},~\mathbf{n}_{i} = \frac{\u_{i} \times \u_{i+1}}{||\u_{i} \times \u_{i+1}||}
\end{align}
where $\mathbf{b}_i$ acts as a negative angle bisector between the vectors $\x_{i-1} - \x_i$ and $\x_{i+1}-\x_{i}$ and $\mathbf{n}_{i}$ is the unit normal vector of that plane.
\subsection{Connection to Independent E(3)-Equivariant Graph Matching Networks (IEGMNs)}
\label{sec:iegmn}
Independent E(3)-Equivariant Graph Matching Networks~\citep{equidock} combine Graph Matching Networks (GMN)~\citep{gmm} and E(3)-Equivariant Graph Neural Networks~\citep{egnn}, to characterize interactions between an input pair of graphs $G_1 = (V_1,E_1), G_2=(V_2,E_2)$. IEGMNs utilize inter and intra-message passing to update the node features and the spatial encodings. We adopt the notation from \cite{equidock}: $m_{ij}$ denotes the messages between nodes $i$ and $j$, $m_{n}$ represents the averaged message over all the neighbors, $\mu_{ij}$ represents the intra-connection edge features, and $a_{ij}$ are the attention coefficients. These features create an aggregated external message in $\mu_{1}$ and $\mu_{2}$. The aggregated external messages are then used to update the node feature embedding $\mathbf{h}_n$, and the spatial embedding $\x_{n}$ for all nodes in both graphs.

As outlined in (Table~\ref{tab:iegmn_tab_app}), \abode~ shares strong similarities with IEGMN. Interestingly, both methods compute two kinds of messages (one kind pertains to messages for nodes of the same type/graph, and the other for a different type/graph). The role of $\mathbf{\mu}_{ij}$ is seem to be played by $\mathbf{m}_{ij}^{\prime,int}$ and $\mathbf{m}_{ij}^{\prime,ext}$ to update the corresponding node and spatial embeddings.
\begin{table*}[!hbt]
\centering
    \caption{\abode~ as a variant of Independent E(3)-Equivariant Graph Matching Network (IEGMN) applied to interactions among two graphs $G_{1} = (V_1,E_1)$ and $G_{2} = (V_2,E_2)$. Here, $e_{ij} \in E_1 \cup E_2$; $n \in V_1 \cup V_2$; $\texttt{RBF}(\x_i,\x_j;\sigma) = \texttt{exp}(-||\x_{i}^{(l)} - \x_{j}^{(l)}||^{2}/\sigma)$; $h_n$ and $x_n$ denote, respectively, the node embedding and the spatial embedding; $a_{ij}$ are attention based coefficients; $\phi^{x}$ is a real-valued (scalar) parametric function; $\phi^{h,e}$ are parametric functions (MLPs); $\textbf{f}_{ij},\textbf{f}_{i}$ are the original edge and node features; $\beta,\eta$ are scaling parameters and $\mathbf{W}$ is a learnable matrix. For \abode~, $\alpha_{i,j}$ are the attention coefficients; $\mathbf{W}_{1},\ldots,\mathbf{W}_{6}$ are learnable weight parameters; $d$ is the hidden size of each head; $\mathcal{N}_{int}(i)$ are the neighbours $j$ of node $i$ such that $k_{ij} = 1$, and $\mathcal{N}_{ext}(i)$ are the neighbours such that $k_{ij} = 2$.  }
    \centering
    \label{tab:iegmn_tab_app}
    \vskip 0.1in
    \resizebox{\textwidth}{!}{
    \begin{tabular}{lcc}
        \toprule
        Method  & IEGMN layer &  \abode   \\
        \midrule
        
        \multirow{2}{4em}{Edge}  & $\mathbf{m}_{i j}=\varphi^e\left(\mathbf{h}_i^{(l)}, \mathbf{h}_j^{(l)}, \operatorname{RBF}\left(\mathbf{x}_i^{(l)}, \mathbf{x}_j^{(l)} ; \sigma\right), \mathbf{f}_{i j}\right)$ & $\alpha_{i, j}=\operatorname{softmax}\left(\frac{(\mathbf{W}_{3}\z_{i})^{\top}(\mathbf{W}_{4}\mathbf{z}_j+\mathbf{W}_{6}\mathbf{e}_{i,j}).}{\sqrt{d}}\right)$  \\
        
         & $\mathrm{m}_n=\frac{1}{|\mathcal{N}(n)|} \sum_{j \in \mathcal{N}(n)} \mathrm{m}_{n j}$  & $m_i^{\prime}=\sum_{j \in \mathcal{N}_i} \alpha_{i, j}\left(\mathbf{W}_2 \mathbf{z}_j+\mathbf{W}_6 \mathbf{e}_{i j}\right)$ \\
         \midrule
        Intra and Inter connections  & $\boldsymbol{\mu}_{i j}=a_{i j} \mathbf{W h}_j^{(l)}, \forall i \in \mathcal{V}_1, j \in \mathcal{V}_2 \text { or } i \in \mathcal{V}_2, j \in \mathcal{V}_1$  &$m_{ij}^{\prime,ext} = \alpha_{i,j}\left(\mathbf{W}_2 \mathbf{z}_j+\mathbf{W}_6 \mathbf{e}_{i j}\right),~m_{ij}^{\prime,int} = \alpha_{i,j}\left(\mathbf{W}_2 \mathbf{z}_j+\mathbf{W}_6 \mathbf{e}_{i j}\right)$   \\
        & $\boldsymbol{\mu}_i=\sum_{j \in \mathcal{V}_2} \boldsymbol{\mu}_{i j}, \forall i \in \mathcal{V}_1, \quad \boldsymbol{\mu}_k=\sum_{l \in \mathcal{V}_1} \boldsymbol{\mu}_{k l}, \forall k \in \mathcal{V}_2$  & $\textbf{m}_i^{\prime,int}=\sum_{j}^{\mathcal{N}_{int}(i)}m_{ij}^{\prime,int},~\textbf{m}_i^{\prime,ext}=\sum_{j}^{\mathcal{N}_{ext}(i)}m_{ij}^{\prime,ext}$ \\
        \midrule
        Node embedding & $\mathbf{h}_n^{(l+1)}=(1-\beta) \cdot \mathbf{h}_n^{(l)}+\beta \cdot \varphi^h\left(\mathbf{h}_n^{(l)}, \mathbf{m}_n, \boldsymbol{\mu}_n, \mathbf{f}_n\right)$ & $\mathrm{a}_i^{\prime}=\mathbf{W}_1 \mathbf{a}_i+\mathbf{m}_i^{\prime, \text{int}}+\mathbf{m}_i^{\prime,\text{ext}}$  \\
        \midrule
        Coordinate embedding & $\mathbf{x}_n^{(l+1)}=\eta \mathbf{x}_n^{(0)}+(1-\eta) \mathbf{x}_n^{(l)}+\sum_{j \in \mathcal{N}(n)}\left(\mathbf{x}_n^{(l)}-\mathbf{x}_j^{(l)}\right) \varphi^x\left(\mathbf{m}_{nj}\right)$ & $\mathrm{s}_i^{\prime}=\mathbf{W}_{1}\s_i+\mathbf{m}_i^{\prime,\text{int}}+\mathbf{m}_i^{\prime, \text{ext}}$  \\
        \bottomrule
    \end{tabular}}
    \vskip -0.1in
\end{table*}
\subsection{Implementation}
\label{sec:implement}
We implemented AbODE in PyTorch \cite{paszke2019pytorch}. We used three layers of Transformer Convolutional Network \cite{shi2020masked} with hidden embedding dimensions of $128-256-64$. The ODE solver operated over time-steps $t \in [0, 200]$, where we took the last time step value as the final prediction of the model. The ODE system is solved with the Adaptive heun solver with an adaptive step size. We train the models for 10000 epochs with the Adam optimizer and use a batch size of 300.

\subsection{ Encoding the non-CDR Antibody sequence} \label{sec:whole_cond}
We encode the sequence and structural information present in the constant regions of the antibody sequences in the heavy chain. We consider the sequences that occur to the left or are before the CDR-H1, H2, and H3 sequences and denote them as $\z_{<i}$ where $i$ is the location of the first CDR-H1, H2,H3 sequence. We used a 2-layer Transformer Convolutional Network with encoding dimensions $64-16$ to encode the features into a 16-dimension encoding vector denoted as $h_{<i}$. The spatial neighborhood is defined as a k-nearest neighbor graph over the spatial domain for residues, where $k=5$ and $\{e_{ij}\}_{j \in \mathcal{N}_{<i}}$ are the corresponding edge features.
\begin{align}
    h_{<i} &= \phi^{enc}(\z_{<i},\z_{\mathcal{N}_{<i}},\{e_{ij}\}_{j \in \mathcal{N}_{<i}}) \\
    h &= \texttt{Agg}(h_{<i})
\end{align}

To have one encoding vector per antibody, we use \texttt{Agg} to obtain an aggregated encoding $h$, which in turn plays a role in dynamics as,
\begin{align}
    \dz_{i}(t) = \frac{\partial \z_{i}(t)}{\partial t} = f_{\psi}(t,\z_{i}(t),\z_{N(i)}(t),\{\mathbf{e}_{ij}(t)\}_{j},h)
\end{align}
The encoding model and the dynamics are trained simultaneously using the loss described in section 3.4, with the same hyperparameters.
\subsection{Fixed Backbone sequence design} \label{sec:backbone_design}
We evaluate our method for de novo protein sequence design that can fold into a given backbone structure, also known as fixed backbone structure design. In addition to the current features, in Eq~\ref{eq:edge_feat}, we utilized the node and edge features listed in \citet{jing2020learning}, which can be derived solely from backbone coordinates and the protein structures are fixed from the beginning. We follow the same initialization for the amino acid labels and due to the large length of protein sequences and memory constraints, we restrict to k = 30 nearest neighbors when defining the spatial neighborhood. Since this task only requires predicting the sequence, so we utilized only the sequence loss $\calL_{seq}$ as defined in Eq.~\ref{eq:entropy} for training in this setting. 

We use the CATH 4.2 dataset by \citet{ingraham2019} where we discard the redundant, NaN coordinates, and follow a similar experimental setting and split as previous works for a fair comparison. We followed the same hyperparameters as used in other cases, with a batch size of 10.


\end{document}